\documentclass{bmvc2k}

\usepackage{marginnote}
\usepackage{algorithm}
\usepackage{algpseudocode}

\newcommand{\betavae}{$\beta$-VAE}
\newcommand{\btcvae}{$\beta$-TCVAE}

\newcommand{\vect}[1]{\boldsymbol{#1}}

\newtheorem{theorem}{Theorem}

\title{Variational Autoencoders with Decremental Information Bottleneck for Disentanglement}

\addauthor{Jiantao Wu}{jiantao.wu@surrey.ac.uk}{12}
\addauthor{Shentong Mo}{shentonm@andrew.cmu.edu}{4}
\addauthor{Xingshen Zhang}{zhxshd@outlook.com}{5}
\addauthor{Muhammad Awais}{muhammad.awais@surrey.ac.uk}{12}
\addauthor{Sara Atito}{sara.atito@surrey.ac.uk}{12}
\addauthor{Zhenghua Feng}{z.feng@surrey.ac.uk}{123}
\addauthor{Lin Wang}{wangplanet@gmail.com}{5}
\addauthor{Xiang Yang}{anson@mintyea.com}{6}

\addinstitution{
 Surrey Institute for People Centred AI,\\
 University of Surrey,\\
 Guildford, UK
}
\addinstitution{
 Centre for Vision Speech and Signal Processing,\\
 University of Surrey,\\
 Guildford, UK
}
\addinstitution{
School of Computer Science and Electronic Engineering,\\
 University of Surrey,\\
 Guildford, UK
}
\addinstitution{
Carnegie Mellon University, \\
Pittsburgh, USA
}
\addinstitution{
 University of Jinan,\\
 Jinan, China
}
\addinstitution{
Mintyea Tech, Inc. \\
Hangzhou, China
}

\runninghead{Jiantao, et al}{DeVAE} 

\begin{document}

\maketitle


\begin{abstract}

One major challenge of disentanglement learning with variational autoencoders is the trade-off between disentanglement and reconstruction fidelity. Previous studies, which increase the information bottleneck during training, tend to lose the constraint of disentanglement, leading to the information diffusion problem. 
In this paper, we present a novel framework for disentangled representation learning, DeVAE, which utilizes hierarchical latent spaces with decreasing information bottlenecks across these spaces. The key innovation of our approach lies in connecting the hierarchical latent spaces through disentanglement-invariant transformations, allowing the sharing of disentanglement properties among spaces while maintaining an acceptable level of reconstruction performance. We demonstrate the effectiveness of DeVAE in achieving a balance between disentanglement and reconstruction through a series of experiments and ablation studies on dSprites and Shapes3D datasets. \href{https://github.com/erow/disentanglement_lib/tree/pytorch#devae}{Code} is available.

\end{abstract}

\section{Introduction}
Unsupervised learning~\citep{wu2018unsupervised} is essential for bridging the gap between human and machine intelligence. 
Disentanglement learning is a promising approach for obtaining explanatory representations from observations without supervision, mimicking human intelligence~\citep{Bengio.2013}. 
Variational autoencoders (VAEs)~\cite{Kingma.2013} are widely used for disentanglement learning, with methods like beta-VAE~\cite{Higgins2017betavae} introducing a penalty (weighted by $\beta$) on the Kullback–Leibler (KL) divergence to promote disentanglement. However, there is a trade-off between disentanglement and reconstruction fidelity in beta-VAE.

To address this trade-off, some methods utilize a dynamic controlling strategy for $\beta$~\cite{Burgess.2018,shao2022rethinking,DBLP:journals/ml/WuDEFT22}. Generally, a high initial $\beta$ value is set to enforce VAEs disentangle at the beginning. Then, the value of $\beta$ is gradually reduced to facilitate reconstruction. Since $\beta$ controls the Information Bottleneck (IB)~\cite{Tishby.1999,Burgess.2018}, these methods are called \textit{incremental VAEs}, where the IB increases during training. As a result, incremental VAEs achieve a good balance by optimizing disentanglement and reconstruction in separate time spans.

In this work, we propose an alternative approach to address the conflict between optimizing disentanglement and reconstruction. Our primary motivation is to optimize disentanglement and reconstruction simultaneously by creating multiple latent spaces. Each latent space focuses on different tasks, either optimizing disentanglement or reconstruction, while our framework ensures these spaces share disentanglement properties. This approach enables simultaneous optimization of both disentanglement and reconstruction.

Specifically, we introduce DeVAE, a VAE framework with hierarchical latent spaces (HiS) that applies a novel IB-decremental strategy and a disentanglement-invariant transform (DiT) operator. DeVAE gradually decreases the information bottleneck across latent spaces, constrains the first space for reconstruction, and learns factors in subsequent spaces using narrow IBs. The disentanglement-invariant transform operator guarantees that the learned disentangled representation is shared among the latent spaces.

Our contributions can be summarized as follows:
\begin{itemize}
    \item We introduce a novel framework, DeVAE, which employs hierarchical latent spaces with decreasing information bottlenecks across the spaces, offering a new approach to balance disentanglement and reconstruction fidelity.
    \item We develop the disentanglement-invariant transformation, a key innovation that connects hierarchical latent spaces and enabling the sharing of disentanglement properties among them while maintaining a high level of reconstruction performance.
    \item We conduct comprehensive experiments and ablation studies on benchmark datasets, i.e.  dSprites and Shapes3D, demonstrating the effectiveness of DeVAE in achieving a balance between disentanglement and reconstruction.
\end{itemize}

\section{Related Work}

\paragraph{Disentanglement Learning.} 
Disentanglement learning aims to learn generative factors existing in the dataset~\cite{Bengio.2013}.
Although the formal definition of disentanglement is still an open topic, it is widely accepted that the redundancy between latent variables diminishes disentanglement~\citep{Do.2020}.
Penalizing the Total Correlation (TC)~\citep{Watanabe.1960} is an important direction in disentanglement learning, and many state-of-the-art (SOTA) methods are based on it~\citep{Chen2018betatcvae}.
Predictability Minimization (PM) algorithm ~\citep{Schmidhuber.1992} promotes factorial codes but only works for binary codes;
Though ICA~\citep{comon1994independent} and PCA~\citep{wold1987principal} ensure independence theoretically, they extract linear representations.
Recently, deep learning has made this more feasible. 
FactorVAE~\citep{Kim2018factorvae} applies an adversarial training method to approximate and penalize the TC term.
\btcvae{}~\citep{Chen2018betatcvae} decomposed the KL term into three parts: mutual information (MI), total correlation (TC), and dimensional-wise KL (DWKL).
They achieve good performance by optimizing the TC term and avoiding penalizing the MI term.
However, the TC-based methods introduce a strong assumption that generative factors are independent, which is impractical for real-world problems.

\paragraph{Information Bottleneck.} 
Information bottleneck theory~\citep{Tishby.1999,Shannon.1948} plays a vital role in interpreting neural networks.
Some methods encourage disentanglement by increasing the IB during training~\citep{Burgess.2018}.
These methods differ in the way they expand the IB. 
CascadeVAE~\citep{JeongS19a} sequentially relieves one latent variable at each stage to increase the IB.
DynamicVAE~\citep{shao2022rethinking} designs a non-linear PI controller for manipulating $\beta$ to control the steadily increasing IB.
DEFT~\citep{DBLP:journals/ml/WuDEFT22} applies a multi-stage training strategy with separated encoders to extract factors separately at different stages.
However, the above incremental models, which increase the IB during training, suffer from the information diffusion (ID) problem~\citep{DBLP:journals/ml/WuDEFT22}, as the disentangled representation may diffuse the learned information into other variables when expanding the IB.

\paragraph{Hierarchical Latent spaces.} 
Normalizing Flow~\citep{rezende2015variational,kingma2016improved} uses hierarchical latent spaces to generate an arbitrary distribution. Unlike Normalizing Flow, each space in our model aims to encourage disentanglement or reconstruction. Additionally, Normalizing Flow gradually increases the complexity of the output distribution after entering a new space. In contrast, our model reduces the complexity space by space.

\section{Methodology}

\subsection{Preliminaries}

\paragraph{Problem Setup \& Notations.}

Disentanglement learning aims to learn the factors of variation which raises the change of observations.
Given a set of samples $\vect{x} \in \mathcal{X}$, they can be uniquely described by a set of ground-truth factors $\vect{c}\in \mathcal{C}$.
Generally, the generation process $g(\cdot)$ is invisible $\vect{x} = g(\vect{c})$.
We say that a representation for factor $\vect{c}_i$ is disentangled if it is invariant for the samples with $\vect{c}_j$.
We use variational inference to learn the disentangled representation for a given problem.
$p(\vect{z}|\vect{x})$ denotes the probability of $\vect{z}=f(\vect{x})$, $p(\vect{x}|\vect{z})$ denotes the probability of $\vect{x} = g(\vect{z})$.
The representation function is a conditional Bayesian network of the form $q_\phi(\vect{z}|\vect{x})$ to estimate $p(\vect{z}|\vect{x})$.
The generative model is another network of the form $p_\theta(\vect{x}|\vect{z})p(\vect{z})$.
$\phi,\theta$ are trainable parameters.

\begin{figure}
    \centering
    \includegraphics[width=.8\linewidth]{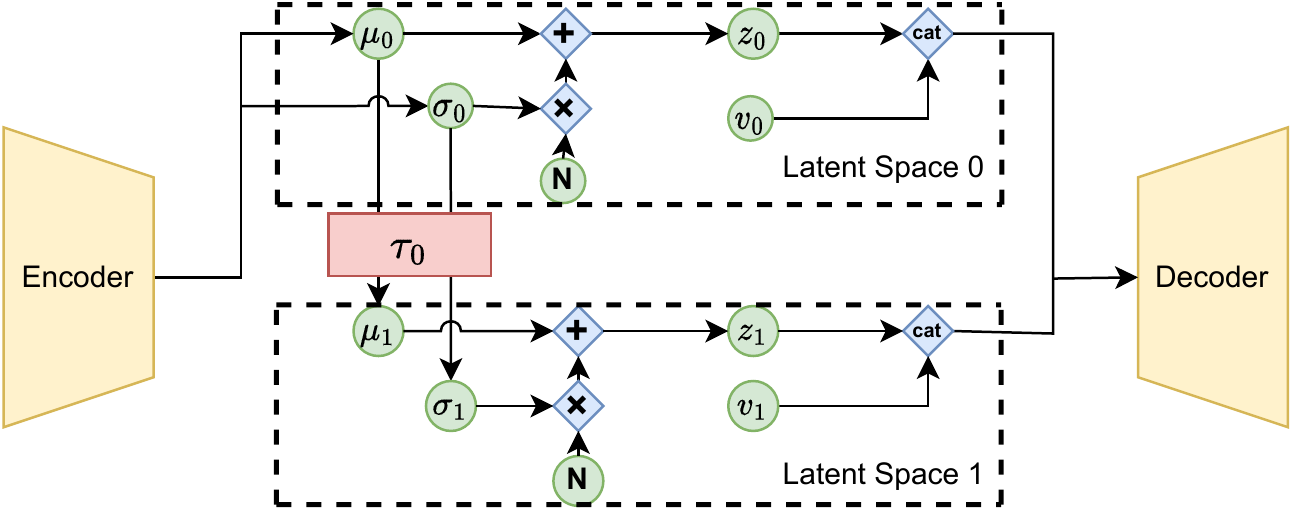}
    \caption{Illustration of our Decremental Variational Autoencoder (DeVAE). Each space has a pressure $\beta_i$ to control the capacity of IB. $\tau_i$ connects two latent spaces. The first space is our main space to represent inputs. The subsequent spaces are minor spaces to improve disentanglement.}
    \label{fig:model}
    \vspace{-0.5em}
\end{figure}

\paragraph{Revisit VAE \& \betavae{}.}

The VAE framework~\citep{Kingma.2013} computes the representation function by introducing $q_\phi(\vect{z}|\vect{x})$ and optimizing the variational lower bound (ELBO).
\betavae{}~\citep{Higgins2017betavae} introduces the hyperparameter $\beta$ to control the IB:
\begin{equation}\label{eq:vae}
    \begin{aligned}
        \mathcal{L}(\theta, \phi) =  \E_{q_\phi(\vect{z}|\vect{x})}[\log{p_\theta (\vect{x}|\vect{z})}]
        - \beta \  D_{\mathrm{KL}}(q_\phi(\vect{x}|\vect{z}) || p(\vect{z})).
    \end{aligned}
\end{equation}

Consider using \betavae{} to learn a representation of the data; the representation will be disentangled but lose information when $\beta$ is large~\citep{Burgess.2018}.
We can set a large $\beta$ to learn a disentangled representation and a small $\beta$ to learn an informative representation. However, \betavae{} suffers a trade-off between disentanglement and reconstruction, which means that $\beta$ can only optimize one of these two goals.

\subsection{Hierarchical Latent Spaces with Decremental Information Bottleneck}\label{sec:dib}

To maintain the disentanglement constraint while optimizing reconstruction fidelity, we introduce a Hierarchical Latent Space (HiS) with $K$ spaces and assign a pressure $\beta_i$ to the $i$-th space $\mathcal{Z}_i$. Each space promotes disentanglement or reconstruction through a suitable value of $\beta$. The objective of the $i$-th space is given by:
\begin{equation}
    \begin{aligned}
        \mathcal{L}_i(\theta, \phi) =  \E_{q_\phi(\vect{z}_i|\vect{x})}[\log{p_\theta (\vect{x}|\vect{z}_i,\vect{v}_i)}]
        - \beta_i D_{\mathrm{KL}}(q_\phi(\vect{z}_i|\vect{x}) || p(\vect{z})),
    \end{aligned}
\end{equation}
where the first space $q_\phi(\vect{z}_0|\vect{x})$ is a conditional Bayesian network, $v_i$ denotes a $K$-D vector to indicate the index of space, and the subsequent spaces can be calculated by:
\begin{equation}\label{eq:latent}
    \begin{aligned}
        q(\vect{z}_{i+1}|\vect{x}) = \tau_i (\vect{z}_{i+1}|\vect{z}_{i}) q(\vect{z}_{i}|\vect{x}), i\neq0,
    \end{aligned}
\end{equation}
where $\tau_i$ denotes a transformation from $\mathcal{Z}_i$ to $\mathcal{Z}_{i+1}$.

According to information theory, information can only decrease during processing. Therefore, we gradually decrease the IB in the sequential spaces, i.e., $\beta_{i+1}>\beta_i$. Typically, we set $\beta_0=1$ to encourage the first space to focus on reconstructing the original inputs. In this way, sequential spaces aim to disentangle factors of variation by setting narrow bottlenecks.

\subsection{Disentanglement-invariant Transformation}\label{sec:dit}
In this part, we discuss the transformation $\tau_i$ which is vital to optimizing disentanglement and reconstruction simultaneously. If the transformation is arbitrary, the spaces will optimize their goal independently. Therefore, we need a mechanism to connect these goals to balance disentanglement and reconstruction in one space. 
To share disentanglement across all latent spaces, we propose a disentanglement-invariant transformation (DiT) denoted as $\tau$:
\begin{equation}
    \vect{\mu}_{i+1} = h(\vect{w}^1_i) \vect{\mu}_i, \quad 
    \vect{\sigma}_{i+1} = h(\vect{w}^2_i) \vect{\sigma}_i,
    \label{eq:tau}
\end{equation}
where $\vect{z}_{i}\sim \mathcal{N}( \vect{\mu}_i, \vect{\sigma}_i)$, $\vect{w}^1_i, \vect{w}^2_i$ are learnable diagonal matrices of the $i$-th space, $h(\vect{w})=e^{\vect{w}} >0$ is an exponential function to make sure the scale values greater than 0.

We prove that scaling the latent space will not change disentanglement in Theorem 1, see proof in Appendix~\ref{sec:proof}.
\begin{theorem}
    $\vect{w} \cdot \vect{z}$ is disentangled if $\vect{z}$ is disentangled, $\vect{w}$ is a diagonal matrix.
\end{theorem}

\subsection{Optimization Algorithm}

According to Equation~\ref{eq:tau}, we derive the parameters of latent variables for $i$-th space:
\begin{equation}
    \begin{aligned}
        \vect{\mu}_i = h(\sum_{j=0}^{{i-1}} w_j^1) \vect{\mu}_0, \quad
        \vect{\sigma}_i = h(\sum_{j=0}^{{i-1}} w_j^2) \vect{\sigma}_0,\quad
        i>0.
    \end{aligned}
\end{equation}

Applying the chain law, we get the $i$-th KL divergence:
\begin{equation}
    D_{{\mathrm{KL}}_i} = \frac{1}{2}(1+
    2\sum_{j=0}^{{i-1}} w_j^2 + 2\log(\vect{\sigma}_0)
    - h(2\sum_{j=0}^{{i-1}} w_j^2) \vect{\sigma}_0^2
    - h(2\sum_{j=0}^{{i-1}} w_j^1) \vect{\mu}_0^2 )
\end{equation}

The final objective of DeVAE is:
\begin{equation}\label{eq:devae}
 \mathcal{L}(\theta, \phi) = \sum_{i=0}^{K-1} \E_{q_\phi(\vect{z}_i|\vect{x})}[\log{p_\theta (\vect{x}|\vect{z}_i,\vect{v}_i)}]
        -  \sum_{i=0}^{K-1}  \beta_i D_{{\mathrm{KL}}_i}.
\end{equation}

In this work, we aim to prove the validity of the proposed HiS with DiT for optimizing disentanglement and reconstruction simultaneously in different latent spaces. The algorithm of our method is shown in Algorithm~1. Figure~\ref{fig:model} illustrates the architecture of DeVAE with two spaces. We set $K=2$ for simplicity, and we find it is effective in practice. The main space applies $\beta_0=1$ to work as a vanilla VAE. We set a high value of $\beta_1$, adjusting according to problems, to encourage disentanglement.


\begin{algorithm}
\caption{DeVAE: Hierarchical Latent Spaces with Decremental Information Bottleneck}
\begin{algorithmic}[1]
\Require Data $\mathcal{D}=\{\vect{x}_n\}_{n=1}^N$, epochs $T$, learning rate $\eta$, pressure parameters $\beta_0=1, \beta_1$
\State Initialize the encoder and decoder networks $\phi$ and $\theta$
\For{$t=1$ to $T$}
    \For{each $\vect{x}$ in $\mathcal{D}$}
        \State Compute $\vect{\mu}_0, \vect{\sigma}_0$ using the encoder network $q_\phi(\vect{z}_0|\vect{x})$
        \State Sample $\vect{z}_0 \sim \mathcal{N}(\vect{\mu}_0, \vect{\sigma}_0)$
        \State Compute $\vect{\mu}_{i+1} = h(\vect{w}^1_i) \vect{\mu}_i,
    \vect{\sigma}_{i+1} = h(\vect{w}^2_i) \vect{\sigma}_i,$ using DiT
        \State Sample $\vect{z}_1 \sim \mathcal{N}(\vect{\mu}_1, \vect{\sigma}_1)$
        \State Compute reconstruction loss $\mathcal{L}_{rec} = \sum_{i=0}^{1} \E_{q_\phi(\vect{z}_i|\vect{x})}[\log{p_\theta (\vect{x}|\vect{z}_i,\vect{v}_i)}]$
        \State Compute KL divergence losses $D_{{\mathrm{KL}}_0}$ and $D_{{\mathrm{KL}}_1}$
        \State Compute total loss $\mathcal{L}(\theta, \phi) = \mathcal{L}_{rec} - \beta_0 D_{{\mathrm{KL}}_0} - \beta_1 D_{{\mathrm{KL}}_1}$
        \State Update $\phi$ and $\theta$ using gradient descent with learning rate $\eta$
    \EndFor
\EndFor
\end{algorithmic}
\end{algorithm}

\section{Experiments}

\subsection{Experimental Setup}

\paragraph{Datasets.}
The experiment section assesses the proposed DeVAE method on two widely-used datasets, dSprites~\citep{dsprites17} and Shapes3D~\citep{3dshapes18}.  
dSprites has 737,280 binary 64 × 64 x 1 images generated from five factors: shape (3), orientation (40), scale (6), position X (32), and position Y (32).
Shapes3D has 480,000 RGB 64 × 64 × 3 images of 3D shapes generated from six factors: floor color (10), wall color (10), object color (10), object size (8), object shape (4), and azimuth (15).

\paragraph{Evaluation Metrics.}
To evaluate the performance of disentanglement, three disentanglement metrics are applied.
\textbf{MIG}~\citep{Chen2018betatcvae}: the mutual information gap between two variables with the highest and the second-highest mutual information.
\textbf{FactorVAE metric}~\citep{Kim2018factorvae}: the error rate of the classifier, which predicts the latent variable with the lowest variance.
\textbf{DCI Dis.}: abbreviation for DCI Disentanglement~\citep{Eastwood.2018}, a matrix of relative importance by regression.
\textbf{Recon.}: abbreviation for Reconstruction Error. We use Squared Error for RGB images (Shapes3D) and Binary Cross Entropy for binary images (dSprites).

\paragraph{Implementation.}
We use a convolutional neural network as the encoder and a deconvolutional neural network as the decoder. 
Detailed architecture can be found in Appendix~\ref{sec:arch}.
The activation function is ReLU\@.
The optimizer is Adam~\citep{DBLP:journals/corr/KingmaB14} with a learning rate of $1e^{-4}$, \(\beta_1 = 0.9,\ \beta_2 = 0.999\).
We employed a large batch size of 256 to accelerate the training process.
All experiments train 300, 000 iterations by default.
For the hyper-parameters, we set $\beta=12$ for  \btcvae{}, $\beta=6$ for \betavae{}, and $K_i=0.001,K_p=0.01$ for DynamicVAE, and $\{\beta_i\}=[1,40]$ for DeVAE.
We set $\beta_0 = 1$ to reconstruct image details and set $\beta_1 = 40$ to filter hard factors (shape, orientation) according to DEFT~\citep{DBLP:journals/ml/WuDEFT22}.

\subsection{Comparison to Prior Work}

To demonstrate the effectiveness of the proposed DeVAE, we compare it to three typical disentanglement methods:
1) $\beta$-VAE~\citep{Higgins2017betavae}: the baseline model for disentanglement and also the special case of DeVAE when $\beta_0=\beta_1$;
2) $\beta$-TCVAE~\citep{Chen2018betatcvae}: the SOTA method for penalizing TC;
3) Dynamic-VAE~\citep{shao2022rethinking}: the SOTA method for incremental VAEs.

\begin{table}[t]

\begin{tabular}{@{}lccccc@{}}
dataset  & model      & MIG & DCI dis. & FactorVAE & Recon. \\ \midrule
dSprites & DeVAE      & 0.34$\pm$ 0.02   & 0.53$\pm$ 0.02     & 0.80$\pm$ 0.03    & 48.31$\pm$ 27.98        \\
 & DynamicVAE & 0.35$\pm$ 0.01   & 0.53$\pm$ 0.01              & 0.82$\pm$ 0.05   & 19.25$\pm$ 1.85         \\
 & $\beta$-TCVAE(12.0)  & 0.29$\pm$ 0.09   & 0.47$\pm$ 0.08     & 0.73$\pm$ 0.08    & 73.04$\pm$ 3.41         \\
 & $\beta$-VAE(6.0)   & 0.17$\pm$ 0.05   & 0.30$\pm$ 0.07     & 0.74$\pm$ 0.05    & 48.75$\pm$ 2.84         \\
\midrule
shapes3D & DeVAE      & 0.53$\pm$ 0.11   & 0.71$\pm$ 0.02     & 0.79$\pm$ 0.02    & 46.81$\pm$ 13.97        \\
 & DynamicVAE & 0.54$\pm$ 0.04   & 0.68$\pm$ 0.03          & 0.87$\pm$ 0.10             & 31.02$\pm$ 3.56         \\
 & $\beta$-TCVAE(12.0)  & 0.49$\pm$ 0.11   & 0.73$\pm$ 0.07     & 0.78$\pm$ 0.01    & 44.53$\pm$ 5.69         \\
 & $\beta$-VAE(6.0)   & 0.42$\pm$ 0.18   & 0.68$\pm$ 0.06     & 0.82$\pm$ 0.06    & 34.95$\pm$ 2.34         \\ 
\bottomrule
\end{tabular}
\vspace{3mm}
\caption{Quantitative benchmarks on dSprites and shapes3D.}\label{tab:quantitative}
\end{table}

\paragraph[]{Disentanglement \& Reconstruction.}\label{sec:benchmark}

In comparison to prior work, DeVAE demonstrates effectiveness in achieving a balance between disentanglement and reconstruction. 
We conducted experiments on dSprites and Shapes3D where each trail was repeated 10 times with different random seeds and evaluated by MIG, FactorVAE, DCI disentanglement, and reconstruction error. We expect higher values for these metrics except recon. 
On the dSprites dataset, DeVAE achieves an average improvement of 12\% in disentanglement compared to $\beta$-TCVAE and 38\% compared to $\beta$-VAE. Furthermore, the reconstruction error is only half of that in $\beta$-TCVAE. The reconstruction drop on shapes3D is not that kind of large, because we use l2 loss instead of Bernoulli loss. DeVAE gains remarkable disentanglement with accepable reconstruction drop. Overall, DeVAE is competitive with Dynamic-VAE and surpasses both $\beta$-TCVAE and $\beta$-VAE.

\begin{figure}
\centering
\includegraphics[width=.95\linewidth]{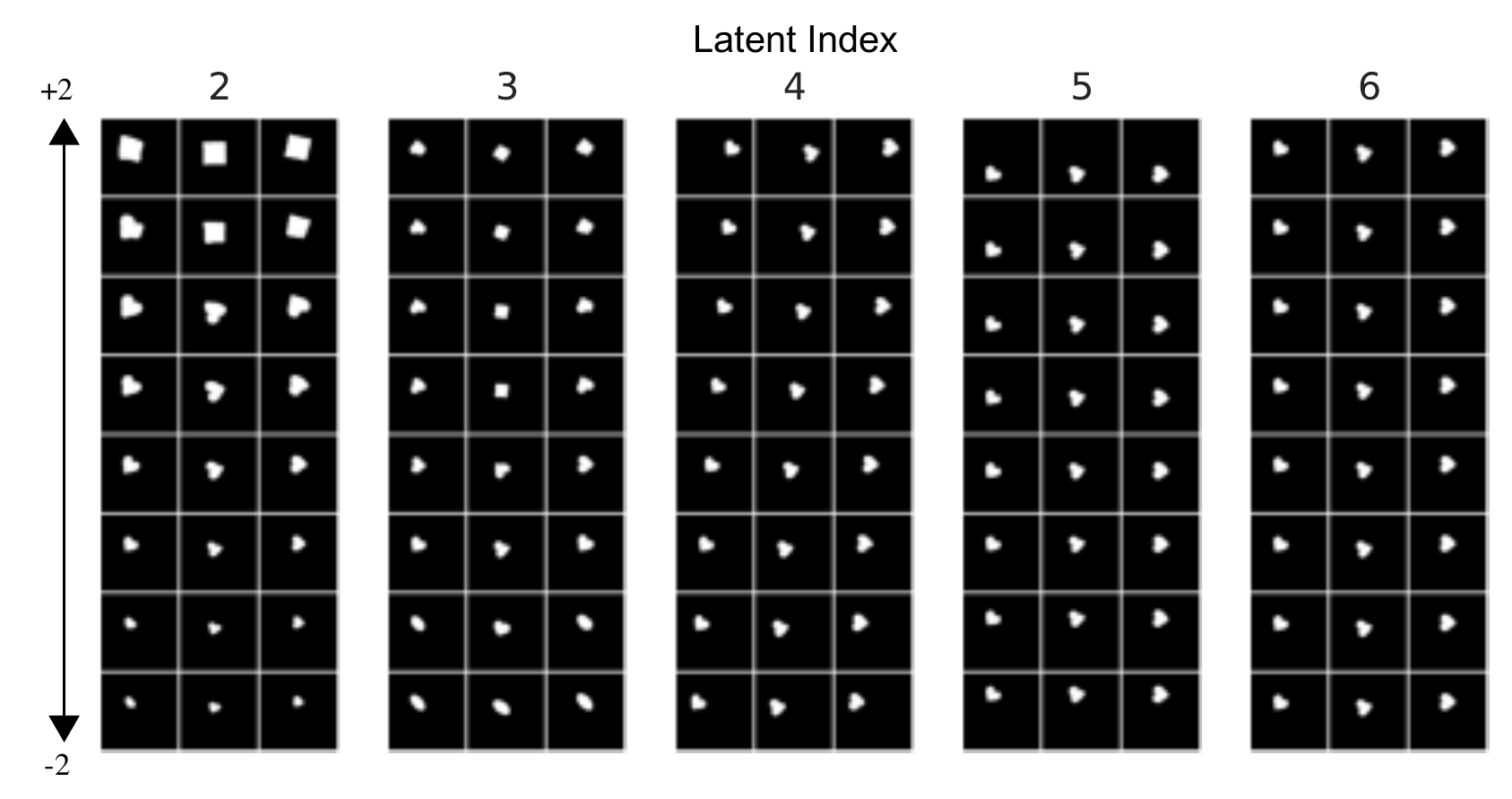}    
\caption[]{Latent traversal on dSprites. Each block shows the generated images of traversing the latent variable (title) from -2 to 2 with three different random sampling.}
\label{fig:traversal}
\end{figure}

\paragraph{Qualitative Visualization.} 
Qualitative analysis is conducted to assess disentanglement by visualizing latent traversals~\cite{Higgins.2018} as shown in Figure~\ref{fig:traversal}. 
Specifically, each row reveals the reconstruction images from one dimension of the latent space  systematically varied from -2 to 2 while keeping the others fixed. For each variation, the decoder of the VAE generates new images with three random seeds.
We choose the top5 dimensions with the highest KL divergence to visualize their latent traversals.
DeVAE successfully disentangles position X and position Y by latent 4 and 5.
The hard factors, shape, scale, and orientation, are still a challenge in this domain.
More examples can be found in the Appendix~\ref{sec:visualization}.

\begin{figure}[t]
    \centering
    \includegraphics[width=.98\linewidth]{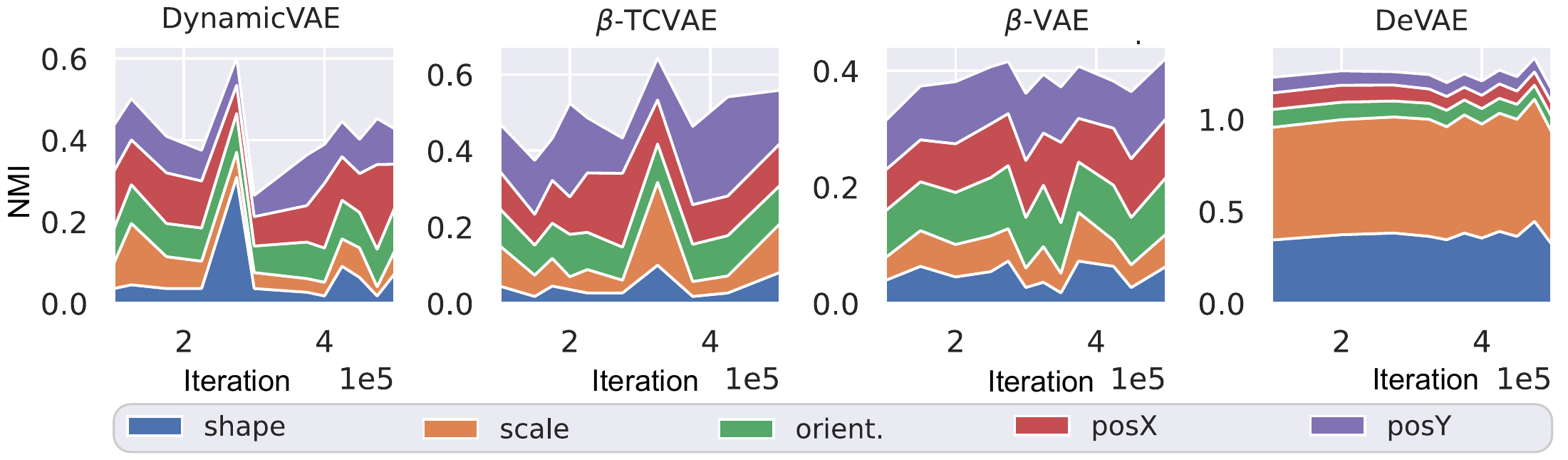}
    \caption{Comparison results of information diffusion. Each colored curve denotes the learned information that belongs to one factor over training iterations.}
    \label{fig:ID}
\end{figure}

\paragraph{Preventing Information Diffusion.}\label{sec:ID}

Information diffusion is a phenomenon where one factor's information diffuses into other latent variables during training, leading to fluctuations in disentanglement scores~\cite{DBLP:journals/ml/WuDEFT22}. 
We argue that our framework can solve the problem effectively due to removing the dynamic controlling strategy.
Figure~\ref{fig:ID} demonstrates the changes in mutual information for the latent variable with the highest KL during training.
NMI refers to the normalized mutual information, calculating the mutual information between one latent variable and one factor divided by the maximum information.
The results show that Dynamic-VAE loses information significantly at iteration 3e5, indicating that the learned structure of representation is destroyed when expanding the information bottleneck (IB). On the other hand, DeVAE demonstrates a relatively steady trend of increasing information, thanks to consistent regularization. DeVAE overcomes the drawbacks of traditional IB-based methods by maintaining the constraint of disentanglement.

\subsection{Experimental Analysis}
\label{sec:ablation}

In this section, we conduct ablation studies to evaluate the benefits of the proposed Hierarchical Latent Spaces (HiS) and Disentanglement-invariant Transformation (DiT). We also explore the effect of these spaces on the balance between disentanglement and reconstruction.

\begin{table}[t]
    \centering
    \begin{tabular}{cccccccccc}
        \multirow{2}{*}{MS}  & \multirow{2}{*}{HiS}    &
        \multirow{2}{*}{DiT} & \multicolumn{3}{c}{MIG} & &\multicolumn{3}{c}{Recon.}   \\
        \cline{4-6} \cline{8-10}
                             &                         &                            & space0        & space1        & space2        & &space0         & space1         & space2         \\
        \midrule
        \xmark               & \xmark                  & \xmark                     & 0.19          & -             & -             & &23.49          & -              & -              \\ 
        \cmark               & \xmark                  & \xmark                     & 0.24          & 0.32          & \textbf{0.35} & &\textbf{22.21} & \textbf{40.79}  & \textbf{62.40} \\ 
        \cmark               & \cmark                  & \xmark                     & 0.24          & 0.29          & 0.30          & &38.82          & 45.48          & 63.78          \\
        \midrule
        \cmark               & \cmark                  & \cmark                     & \textbf{0.35} & \textbf{0.35} & \textbf{0.35} & &43.29          & 75.11          & 175.99         \\ 
        \bottomrule
    \end{tabular}
    \vspace{3mm}
     \caption{Ablation Study on Multiple Space (MS), Hierarchical Structure (HiS) and Disentanglement-invariant Transformation (DiT). }
    \label{tab:ab_component}
\end{table}

\paragraph{HiS \& DiT.}
To demonstrate the effectiveness of the proposed Hierarchy Latent Spaces (HiS) and Disentanglement-invariant Transformation (DiT), we performed ablation experiments on the following scenarios:
1) HiS and DiT are removed, which equals to $\beta$-VAE;
2) HiS is replaced with multiple symmetric encoders instead of the hierarchy encoder, where latent spaces are independent;
3) DiT is replaced with Linear Transformation ($\tau_i(\vect{z_{i+1}}|\vect{z_{i}})=\vect{w} \vect{z_{i}})$, where $\vect{w}$ is an arbitrary matrix.
4) The proposed model DeVAE.
Unlike previous experiments, we compared these models on the dSprites dataset using three spaces ($\{\beta_i\} = [1,10,40]$) to show how DiT affects the connection between spaces.
Table~\ref{tab:ab_component} shows the MIG and reconstruction for each space. 
From the results, we can see that MS and HiS without DiT improve disentanglement slightly. Adding DiT can make sure all latent spaces have same disentanglement. 
DeVAE achieves the best balance through sharing disentanglement at the third space and learning reconstruction at the first space. Thus, the key to DeVAE lies in connecting HiS through DiT.

\begin{figure}
    \centering
    \includegraphics[width=0.9\linewidth]{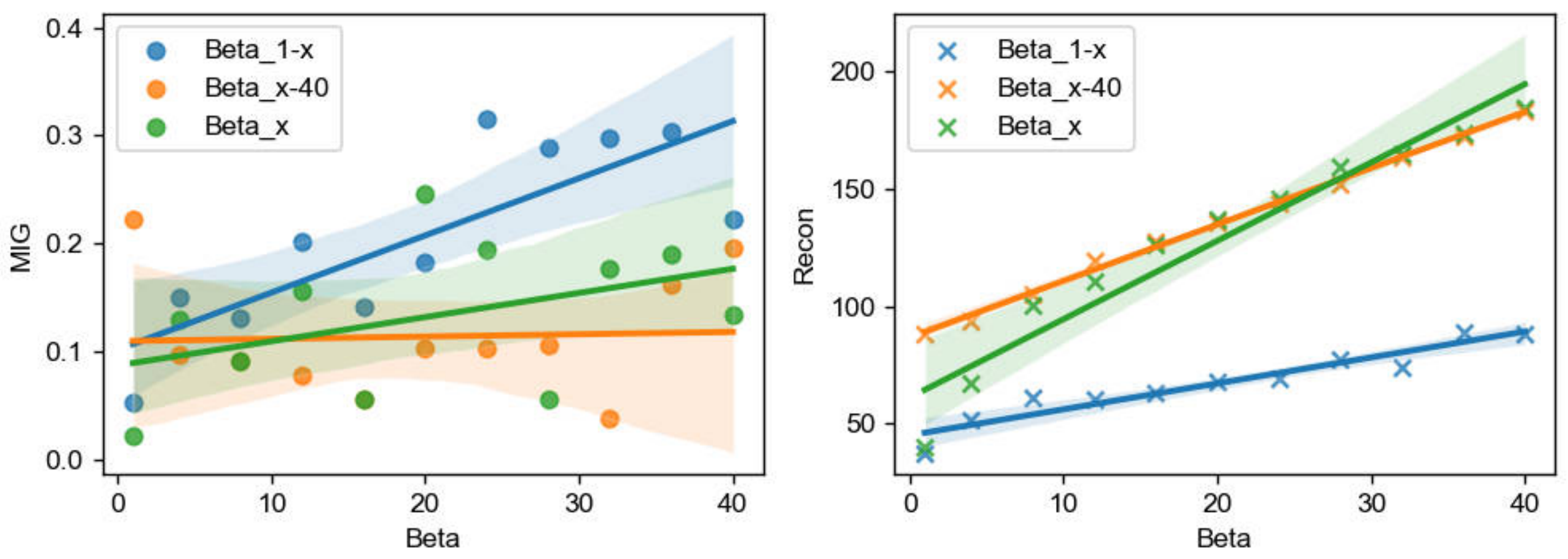}
    \caption{The effects of $\beta_i$ on latent spaces.}
    \label{fig:recon_mig}
\end{figure}
\paragraph{Pressure on Space.}
We argue that the primary role of the first space is to optimize reconstruction and the second space is to optimize disentanglement.
We investigated DeVAE with two latent spaces and applied the following rules to increase beta: 
1) Beta\_x: both spaces apply the same $\beta$, which equals to $\beta$-VAE.
2) Beta\_1-x: only change the pressure of the second space.
3) Beta\_x-40: only change the pressure of the first space.
Figure~\ref{fig:recon_mig} demonstrates the MIG and reconstruction by increasing beta. Each point denotes one experiment with corresponding beta.
One can see that the DeVAE has few reconstruction drop to get a high MIG score.
$\beta_0$ and $\beta_1$ have strong positive correlations with reconstruction error and MIG score respectively, meanwhile, the relationships to MIG and reconstruction are weaker.
Therefore, $\beta_0$ controls reconstruction and $\beta_1$ promotes disentanglement. 


\begin{table}[t]
\centering
\begin{tabular}{llccc}
    \toprule
    Dataset & betas          & MIG                    & Recon.                     & Runtime (min)\\
    \midrule
    dSprites   & [1, 10, 20, 40, 80]        & 0.30$\pm$0.03          & 79.65$\pm$16.06         & 134\\
               & [1, 10, 40]        & \textbf{0.35$\pm$0.02} & 51.99$\pm$26.99              &  109\\
               &  [1, 10]        & 0.16$\pm$0.11          & \textbf{38.19$\pm$02.35}        &101\\
    \midrule
    Shapes3D   & [1, 10, 20, 40, 80]        & 0.53$\pm$0.07          & 70.93$\pm$24.98      &144\\
               & [1, 10, 40]        & \textbf{0.56$\pm$0.01} & 56.09$\pm$4.39              &119 \\
               &  [1, 10]        & 0.55$\pm$0.04          & \textbf{41.43$\pm$5.89}        &103\\
    \bottomrule
\end{tabular}
\vspace{3mm}
\caption{The effect of redundant spaces. }\label{tab:betas}
\end{table}

%

\paragraph{Increasing Spaces.}
The number of spaces is a crucial hyperparameter in our framework. Although the setting $K=2$ achieves remarkable performance, increasing the number of spaces may provide more opportunities to find an optimal solution. However, more spaces require additional computational resources and make it more challenging to optimize the neural network.
In Table~\ref{tab:betas}, we compared tree settings: $\{\beta_i\} =[1,10,20,40,80]$, $\{\beta_i\} =[1,10,40]$, $\{\beta_i\} =[1,10]$.
Fortunately, redundant betas slightly reduce the performance, which means we can create redundant latent spaces spanning a wide range of $\beta$ values to obtain a good model without tuning the hyperparameter extensively.



\section{Conclusion}
In this paper, we propose a novel framework featuring hierarchical latent spaces, where the information bottleneck decreases across spaces. 
These latent spaces are connected through disentanglement-invariant transformations which are the key components to sharing disentanglement among the spaces.
Unlike incremental methods that optimize disentanglement and reconstruction in separate time spans, our work offers insights into optimizing these objectives simultaneously in hierarchical latent spaces. As an original contribution, we have demonstrated how to decouple the two goals, disentanglement and reconstruction, into different latent spaces.

\paragraph{Limitation.}
One limitation of the hierarchical latent spaces is the degradation of reconstruction, which occurs because these spaces are connected and share certain properties, such as disentanglement. Therefore, it is highly desirable to develop a better transformation between latent spaces that results in lower degradation. Future research could focus on improving this aspect of the model to further enhance the balance between disentanglement and reconstruction performance.


\bibliography{reference}

\newpage

\appendix

\section{Appendix}

\subsection{Architecture}\label{sec:arch}
\begin{table}[h]
    
    \centering
    \begin{tabular}{lll}
        \toprule
         {Encoder} & {Decoder} \\
        \midrule
         \(4 \times 4\) conv. 32 stride 2 & FC.\@256 \\
        \midrule
         \(4 \times 4\) conv. 32 stride 2 & FC.\@ \(4 \times 4 \times 64 \) \\
        \midrule
         \(4 \times 4\) conv. 64 stride 2 & \(4 \times 4\) deconv. 64 stride 2 \\
        \midrule
        \(4 \times 4\) conv. 64 stride 2 & \(4 \times 4\) deconv. 32 stride 2 \\
        \midrule
         FC.\@ 256 & \(4 \times 4\) deconv. 32 stride 2\\
        \midrule
         FC.\@ 20 & \(4 \times 4\) deconv. \( c \) stride 2 \\
        \bottomrule
    \end{tabular}
    \vspace{3mm}
    \caption{The architecture details. ``FC.'' denotes fully connected layer, ``conv.'' denotes convolutional layer, ``deconv'' denotes transposed convolution layer. $c$ is the dimension of color channel.}
    \label{tab:architecture}
\end{table}
We use symmetric convolutional networks for encoders and decoders as shown in Table~\ref{tab:architecture}. $c=1$ for dSprites, and $c=3$ for Shapes3D. All layers are activated by ReLU. The final layer of encoder generates 10 variables for \textit{mean} and 10 variables for the \textit{logvar}.

\subsection{Disentanglement-invariant Representations}
\label{sec:proof}
In this section, we prove the proposed disentanglement-invariant transformation.
Consider that we have a new representation by multiplying a diagonal matrix: $\vect{z}' = \vect{w} \vect{z}$, $\vect{w}$.
We can calculate the Covariance between any two latent variables:
\begin{equation}
\begin{aligned}
\operatorname{Cov}(\vect{w}_i \vect{z}_i, \vect{w}_j \vect{z}_j) &=
\mathbb{E}[(\vect{w}_i \vect{z}_i-\mathbb{E}[\vect{w}_i \vect{z}_i])(\vect{w}_j \vect{z}_j-\mathbb{E}[\vect{w}_j \vect{z}_j])] \\
&=\vect{w}_i \vect{w}_j (\mathbb{E}[\vect{z}_j]-\mathbb{E}[ \vect{z}_i] \mathbb{E}[\vect{z}_j]) \\
&=\vect{w}_i \vect{w}_j \operatorname{Cov}(\vect{z}_i,\vect{z}_j),
\end{aligned}
\end{equation}
where the subscript denotes the index of latent variables.
Note that we use a different notion in this section to simplify the formula.

Then we can get the correlation coefficient by
\begin{equation}
\begin{aligned}
    \rho(\vect{w}_i \vect{z}_i, \vect{w}_j \vect{z}_j)&=\frac{\operatorname{Cov}(\vect{w}_i \vect{z}_i, \vect{w}_j \vect{z}_j)}{\sqrt{\operatorname{Var}[\vect{w}_i \vect{z}_i] \operatorname{Var}[\vect{w}_j \vect{z}_j]}}\\
    &= \rho( \vect{z}_i, \vect{z}_j).
\end{aligned}
\end{equation}

Therefore, the correlation matrix will not change by multiplying a diagonal matrix $w, w\neq0$.
The proposed transformation is disentanglement-invariant.

\subsection{ Estimation of $I(\vect{z}_j;\vect{c}_i)$}

Given an inference network $q(\vect{z}|\vect{x})$, we use the Markov chain Monte Carlo (MCMC) method to get samples from $q(\vect{z})$ by the formula $q(\vect{z}) = q(\vect{z|x})p(\vect{x})$.
We use 10, 000 points to estimate $q(\vect{z})$.
Then, we discretize these latent variables by a histogram with 20 bins. 
After discretizing one latent variable, we call a discrete mutual information estimation algorithm to calculate $I(\vect{w}_j \vect{z}_j;\vect{c}_i)$ by a 2D histogram.

\begin{figure}[t]
    \centering
    \includegraphics[width=0.95\textwidth]{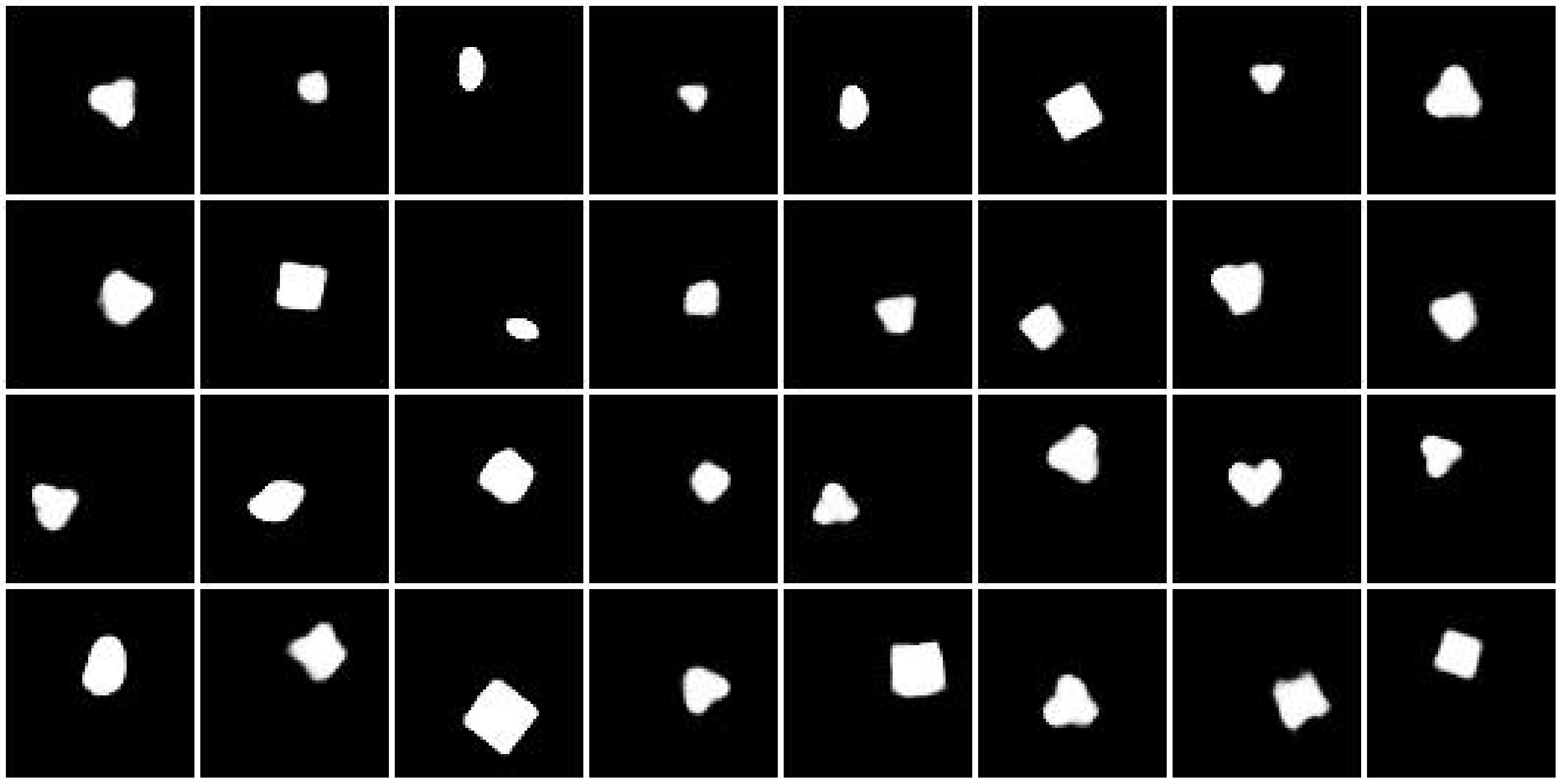}
    \caption{Reconstruction from noise.}
    \label{fig:sampling}
\end{figure}

\subsection{Visualization}\label{sec:visualization}

\begin{figure}[t]
    \centering
    \includegraphics[width=\linewidth]{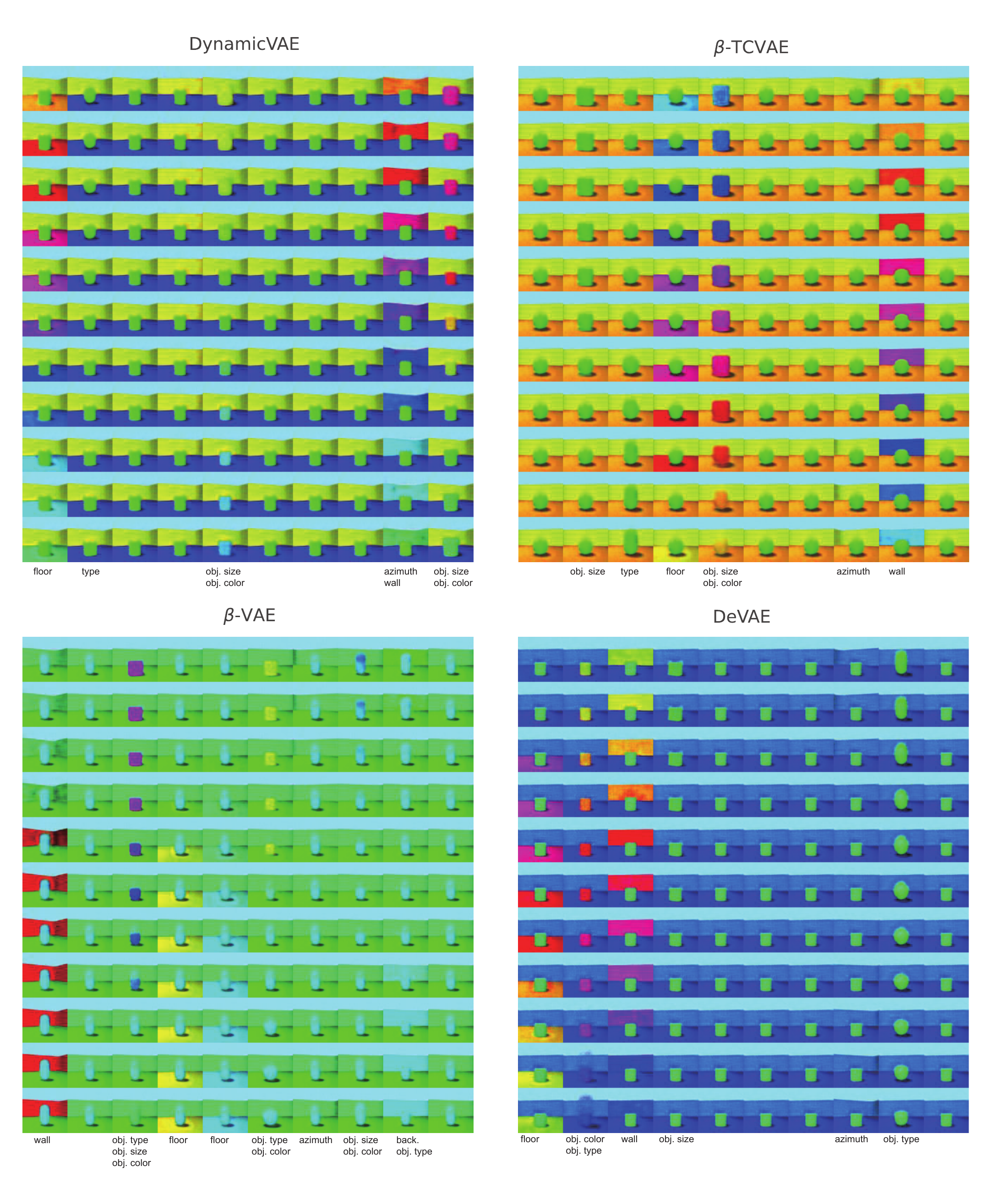}    
    \caption[]{Latent traversal on Shapes3D. ''back.`` denotes background color, ``floor'' denotes floor color, ``obj.'' denotes object, and ``wall'' denotes wall color.}
    \label{fig:traversal_shapes}
\end{figure}

\begin{figure}[t]
    \centering
    \includegraphics[width=\linewidth]{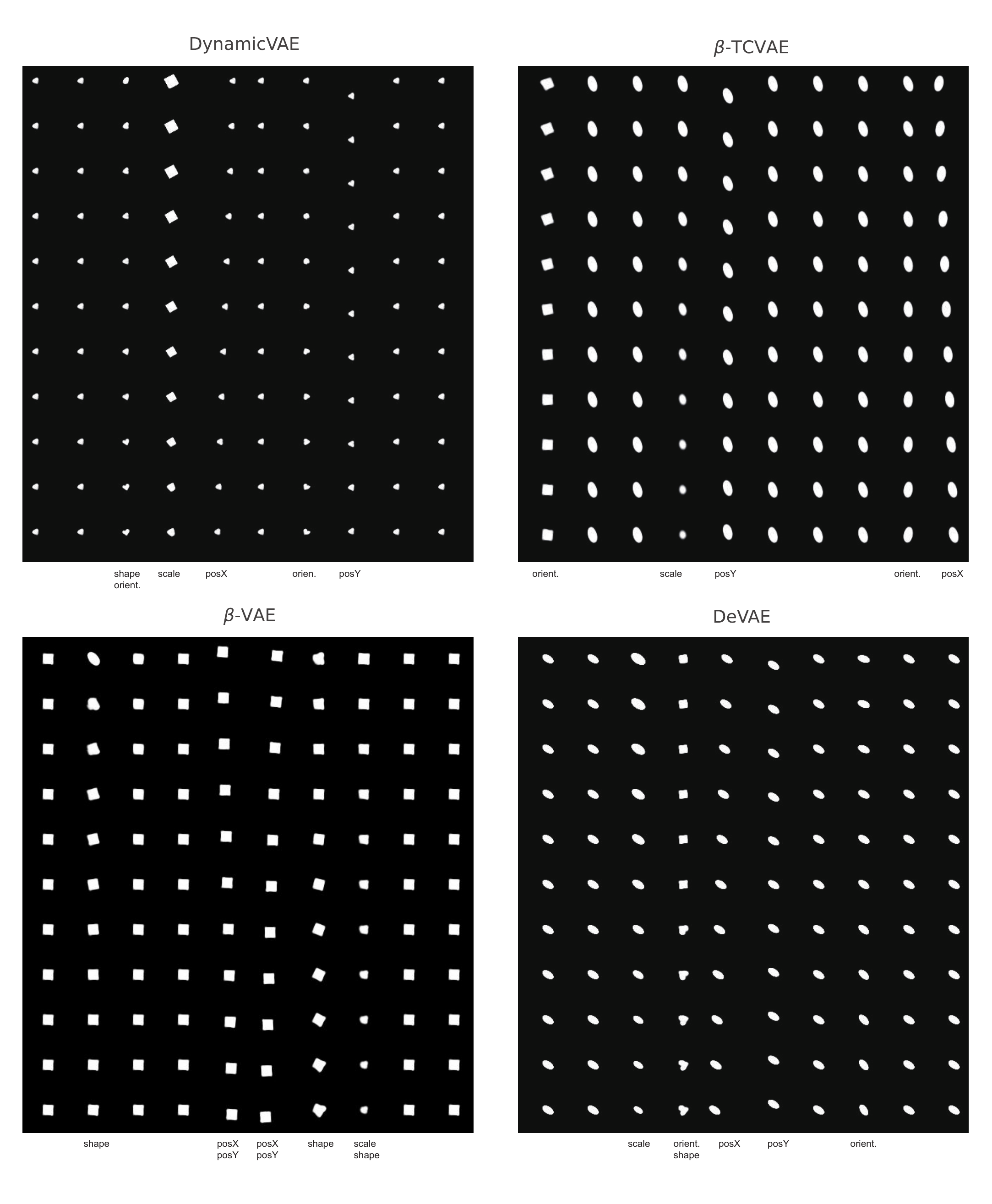}    
    \caption[]{Latent traversal on dSprites.}
    \label{fig:traversal_dsprites}
\end{figure}

\noindent{}\textbf{Latent Traversal.} \quad
We compare DeVAE to others with latent traversals on Shapes3D and dSprites.
Each column shows the generated images by traversing one latent variable from -2 to 2.
From Figure~\ref{fig:traversal_shapes} and Figure~\ref{fig:traversal_dsprites}, we can see that DeVAE has a lower entanglement level. Note that only DeVAE disentangles object size isolated on Shapes3D.

\noindent{}\textbf{Random Sampling.} \quad
We random sample noise from Guassian distribution $\mathcal{N}(0,1)$ and generate images from our disentanglement model trained on dSprites.
As shown in Figure~\ref{fig:sampling}, our model, generating heart, has a high reconstruction fidelity


\subsection{CelebA}

We further conduct experiments on a real dataset CelebA~\cite{liu2015faceattributes}.

\end{document}